\documentclass{article}
\usepackage{natbib}
     \usepackage[final]{neurips_2020_ml4ps}

\usepackage[utf8]{inputenc} 
\usepackage[T1]{fontenc}    
\usepackage{hyperref}       
\usepackage{url}            
\usepackage{booktabs}       
\usepackage{amsfonts}       
\usepackage{nicefrac}       
\usepackage{microtype}      
\usepackage{physics}
\usepackage{graphicx}
\usepackage{leftidx}
\usepackage{cite}
\UseRawInputEncoding

\title{A Quantum-Inspired Probabilistic Model\\ 
for the Inverse Design of Meta-Structures}

\author{Yingtao Luo \\
  University of Washington\\
  Seattle, WA 98195 \\
  \texttt{yl3851@uw.edu} \\
  \And
  Xuefeng Zhu \\
  Huazhong University of Science and Technology \\
  Wuhan, China, 430074 \\
  \texttt{xfzhu@hust.edu.cn} \\
}

\begin{document}

\maketitle

\begin{abstract}
  In quantum mechanics, a norm squared wave function can be interpreted as the probability density that describes the likelihood of a particle to be measured in a given position or momentum. This statistical property is at the core of the microcosmos. Meanwhile, machine learning inverse design of materials raised intensive attention, resulting in various intelligent systems for matter engineering. Here, inspired by quantum theory, we propose a probabilistic deep learning paradigm for the inverse design of functional meta-structures. Our probability-density-based neural network (PDN) can accurately capture all plausible meta-structures to meet the desired performances. Local maxima in probability density distribution correspond to the most likely candidates. We verify this approach by designing multiple meta-structures for each targeted transmission spectrum to enrich design choices.
\end{abstract}

\section{Introduction}

Discovering tailored materials from the starting point of a particular desired functionality, known as \emph{inverse design}, is a rapidly growing field benefitting from the powerful representation learning of machine learning. As a machine learning application in physical sciences, automatic intelligent design is speeding up the discovery of many new materials and refreshing scientists' understanding. For example, previous literatures have reported machine learning applications in discovering chemical reactions, drugs, molecules, nanophotonics and metamaterials. \citet{segler2018} \citet{sanchez2018inverse} \citet{schneider2018automating} Before machine learning, materials were purposefully designed by following physical guiding rules, which belongs to the paradigm of forward design. For metamaterials with negative refractive indices, we tailored structures with locally resonant cells by matching the frequencies of monopole and dipole resonances. \citet{liu2000} To design specific functional metasurfaces, we exploited local resonances to manually engineer phases of the reflected or transmitted sound, which are generally slow and trial-and-error. \citet{ma2016acoustic}

Machine learning with its powerful representation learning opens a door for automatic fast design. \citet{molesky2018} Previous literatures on meta-device design extensively applied machine learning, for example, proposed tandem neural network (TNN) that pre-trains a forward network and then uses the pre-trained module to post-train the final inverse network. \citet{liu2018} This technique can ensure the convergence of loss, but it limits the choice of inverse designs by making a concession of reducing the one-to-many relation into one-to-one. The choice of inverse designs is desirable in real-world scenarios, as some predicted materials may not exist or be hard to be built. While a one-to-one modeling representing the traditional inverse design methods may optimize to an inoperable material structure in real practice, we are interested in enriching our design choices. In other interesting works, deep generative models are leveraged to model the one-to-many function to solve that problem, but their convergence is unstable. \citet{ma2019probabilistic} \citet{liu2018generative} \citet{gulrajani2017}

Above all, the motivation here is to propose an inverse design paradigm to efficiently approximate the multivalued function that governs inverse design. Quantum mechanics reveals a world consisting of pure statistics until observation occurs and the system decoherences. This property has inspired previous literatures to study the quantum-inspired version of evolutionary algorithms \citet{han2002quantum}, recommendation systems \citet{tang2019quantum}, language models \citet{Sordoni2013}, information retrieval models \citet{Peng2018}, and etc. Likewise, it is reasonable to model the quality-factor probabilities of meta-structures as observed states represented by density matrix, where the quality factor represents the likelihood of the designed structure to be the on-demand version. In this work, we propose a quantum-inspired probabilistic density network (PDN) that leverages the quantum-like interference to store the multivalued information in a mixed system. We demonstrate the effectiveness of this approach by retrieving the best fitting meta-structures for a targeted sound transmission over a wide frequency spectrum, with experiments unequivocally demonstrating the effectiveness.

\section{Probabilistic Density Network}
In quantum theory, Dirac's notation is widely used to denote a unit vector as a \emph{ket} $\ket{\psi}$ and its transpose form \emph{bra} $\bra{\psi}$. The outer product of a state is denoted as  $\ket{u}\bra{u}$, representing the projector or the density on the state. To represent the probability of a system, a density matrix is denoted as
\begin{equation}
\rho=\sum_{i}p_i\ket{\psi_i}\bra{\psi_i}
\end{equation}
For a system of pure states, $\rho$ is symmetric, positive semidefinite, and $\tr(\rho)=1$. The diagonal entries give the probability of each state and the off-diagonal elements represent quantum coherences. The joint distribution of multiple systems $S_1, S_2, ..., S_n$, representing the multiple design choices, can be written as a joint density matrix $\rho_{S}=\rho_{S_1}\otimes\rho_{S_2}\otimes...\otimes\rho_{S_n}$. To recover the marginal density matrix, we can use the partial trace operation over the other particles to attain the state of the desired particle:
\begin{equation}
\rho_{S_1}=tr_{S_{2,...,n}}(\rho_{S})=\sum_{j=1}^{n}\leftidx{_{S_{2,...,n}}}{\expval{\rho_{S}}{j}}{_{S_{2,...,n}}}
\end{equation}

In the inverse design scheme, we can leverage this property of quantum computation to train a joint distribution for multiple results and inference any single result via partial trace operation. Recall that in classical scheme, we optimize the conditional probability for representing physical relation
\begin{equation}
L(\theta|X,Y)=\prod_{i=1}^np_{model}(y_i|x_i;\theta)=\sum_{i=1}^nlogp_{model}(y_i|x_i)
\end{equation}

\begin{equation}
argmax_{\theta}\sum_{i=1}^nlog(1/(\sigma\sqrt{2\pi})e^{-(y_i-\hat{y_i})^2/(2\sigma^2)})=argmin_\theta\sum_{i=1}^n(y_i-\hat{y_i})^2
\end{equation}

In a regression task, maximizing likelihood function is equivalent to minimizing \emph{mean square error} loss, as shown in Eq (4), if the target $y_i$ obeys the Gaussian distribution $N(\omega^Tx_i,\sigma^2)$. However, in a multivalued regession, this assumption is no longer satisfied to accurately discover the physical relation. Here, we can train a joint distribution and then obtain every single result by partial trace operation to avoid the dilemma, as shown in Eq (2). Since optimizing high-dimensional diagnol matrices in a classical computer can be computationally costly, here we present a classical probabilistic model as an alternative implementation. The proposed framework has two modules that combines neural model with probability sampling, as shown in Fig 1. In this hybrid architecture, the front end is a neural network that maps a desired transmission spectrum to the parameters of individual Gaussian distributions, other than giving the outputs directly. The rear end uses these parameters to construct a mixed probability by linear superposition, and then sample the output with respect to the probability of each superposed distribution for the meta-structure design. The amplitude of mixed probability can be deciphered as the global wave function of plausible meta-structures. We can simulate the density operator of a system by Gaussian distribution under the Central Limit Theorems 
\begin{equation}
\rho=p(y|x,\theta)=\sum_{i=1}^m\pi_i(x)D(y,\mu_i(x),\sigma_i(x)),\qquad\sum_{i=1}^m\pi_i(x)=1
\end{equation}
To implement this classical version model, an attention-only framework is recommended since it allows adaptive input/output sizes and most materials or molecules have variable lengths or geometries. Here, by adopting our framework, we allow various targets to obey various centralized distributions.
\begin{figure}
  \centering
  \includegraphics[width=12.5cm, height=6.25cm]{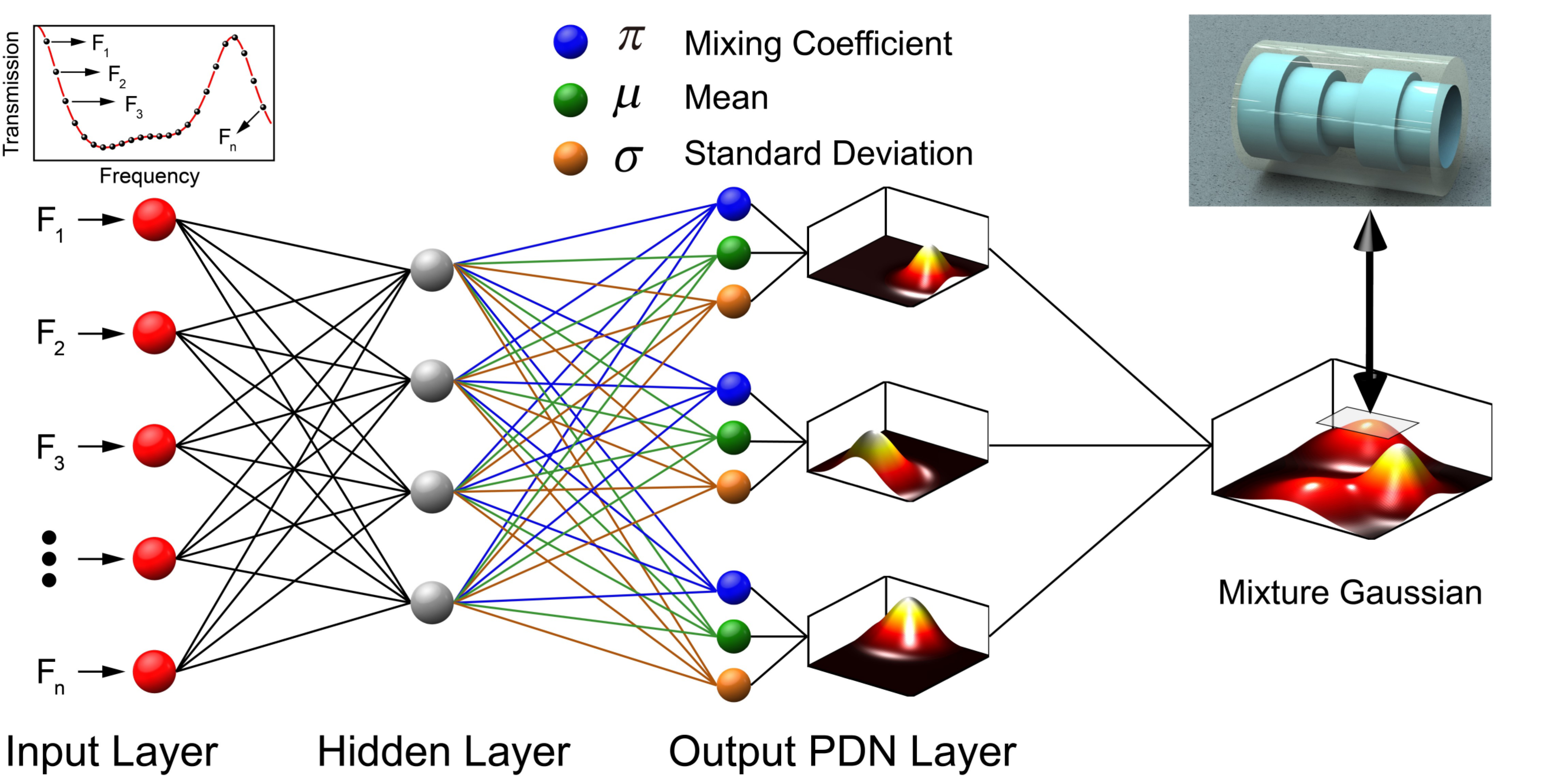}
  \caption{Architecture of the proposed probabilistic density network. The PDN input is the targeted transmission spectrum, while the output is a mixed probability that provides a probabilistic decoherence to derive a plausible meta-structure with similar transmission. Here the mixed probability is superposed by individual probabilities in the output PDN layer, characterized by the parameters of mixing coefficient, mean, and deviation. The local maximum in the mixture Gaussian is mapping to an inversely designed structures with transmission spectra mostly close to the target one.}
\end{figure}

\section{Experimental demonstration}
To give a demonstration, we employ the proposed PDN to inversely design variable cross-section meta-structures based on the target transmission spectrum in acoustics. To collect the labelled data, we utilize a commercial finite element solver COMSOL Multiphysics 5.3TM that is linked to MATLAB to consider the thermo-acoustic effect. Specifically, when the meta-structure has 5 cylindrical layers with radius for each layer sampled from 8 values, we will end up with a number of training data being $8^5$, that is 32768, with the sampled radii at each layer to be either 1.8125 mm, 3.625 mm, 5.4375 mm, 7.25 mm, 9.0625 mm, 10.875mm, 12.6875mm, or 14.5mm. We use uniform sampling to avoid bias in the solution space, but random sampling is also fine. In this work, the input of PDN has 250 dimensions, corresponding to transmittances at frequencies from 20 Hz to 5000 Hz with an interval of 20 Hz, while the output has only 5 dimensions of the radii of the five cylindrical layers in meta-structures. We also randomly sampled 1000 test data from the continuous range of structural parameters without overlap with training data. Here the reverse design of a 5-layer meta-structure is a representative case for testing the model's multivalued inverse design capability. 

We explore with acoustic experiments to verify the effectiveness of PDN model. We firstly employed the labelled dataset to train the PDN model and then fixed the weights for inference. In this example, the target transmission spectrum is featured with a wide bandgap in the frequency range from 20 Hz to 5000 Hz. With the target spectrum as the input, the PDN model outputted a mixed distribution, as shown in Fig. 2(a). Here to visualize the result, we utilize the technique of principal component analysis to reduce data dimensions from 5 to 2 (x and y). With the probability density, we can evaluate the quality factor that characterizes the likelihood of the predicted meta-structure to fulfill the input target in advance. Since the local maxima correspond to the most confident samples, we chose to directly sample at the peaks in Fig. 2(a), as marked by the arrows. In Figs 2(b) and (c), the predicted outputs at the locally highest confidence are located at A1, A2 and A3 in the reduced 2D space, mapping to (14.29, 12.31, 10.10, 1.89, 7.85) mm, (13.60, 7.13, 1.90, 11.38, 12.45) mm and (11.23, 2.05, 6.95, 11.67, 13.66) mm in the full 5D space. As shown in Fig. 2(c), for the three sampled structures, the predicted transmission spectra (blue solid line) are consistent with the target one (red dashed line) and the experimentally measured one (solid spheres).

\begin{figure}
  \centering
  \includegraphics[width=12.5cm, height=6.8cm]{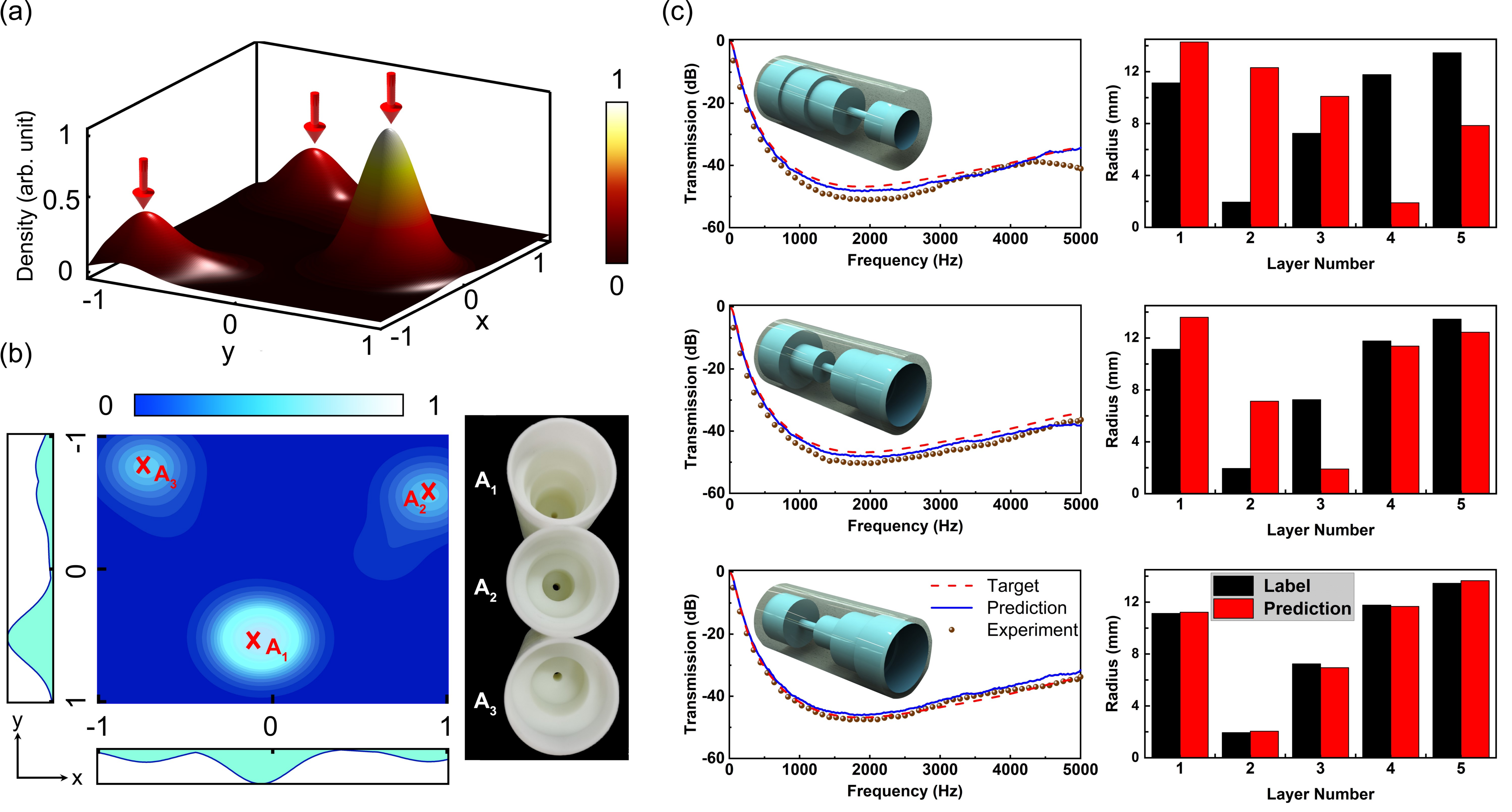}
  \caption{Example of the PDN-based inverse design and experimental demonstration. (a) The output mixture Gaussian for a target transmission spectrum, which is visualized in a 2D plot by reducing the data dimensions from 5 to 2 via the principal component analysis. (b) Exact positions of the local maxima visualized in the contour diagram, where each maximum is mapped to a meta-structure. The local maxima and the corresponding meta-structures are labelled as A1, A2, and A3, respectively. (c) The comparison among the target transmission spectra, the PDN predicted and the experimentally measured ones for the three different meta-structures A1, A2, and A3 from the top down. Each predicted A1, A2, and A3 has different radii per layer, but all get very similar desired transmissions to the target ones. The radii of cylindrical air channels in the labelled structure and the predicted structures are appended on the right-side for reference.}
\end{figure}

\section{Conclusion}
In summary, we have demonstrated a probability-density-based deep learning approach, i.e., PDN, which can solve the multivalued inverse-design problems for implementing physically realizable target functionalities in meta-structures with high accuracy. In acoustics, but not confined to this field, we have successfully employed the PDN to evaluate all the plausible meta-structures for different target transmission spectra. The output of PDN is a joint probability density, simulating the quantum physics, for which the amplitude characterizes how closely the meta-structure fits the desired functionalities. To verify the predictions from PDN, we design meta-structures corresponding to the local maxima in probability density distributions for experimental demonstrations. The measured transmission spectra agree well with both the target and the predicted ones from PDN. The proposed PDN is scalable and unparalleled in the aspect of multivalued inverse-design, which paves the way for fast design of novel devices with complex functionalities by optimizing the geometries.

\section{Potential Impact and Future Vision}
The stably multivalued inverse design is instructive in various subjects and applications, such as the data-driven molecule design, material design, industrial process optimization, chemical reaction path planning, equation solving with multiple solutions, and etc. In the area of metamaterials alone, we are now using the paradigm to help guide the on-demand design of optical and acoustical devices with various functionalities such as low-band adsorption, multi-path sound guiding, and etc. Currently, the inverse design result of a machine learning system does not obey physical constrains, thus the inverse design result is not foolproof yet. For example, the folding and sequence possibilities of a protein structure are endless; some are unable to be tested or synthesized, or are unstable to exist in nature. While the current machine learning systems cannot couple physics, enriching the design choices via the multivalued paradigm can help avoid bad design. In the future, a machine learning model incorporating physics constrain can be an alternative approach.

\small
\bibliography{ref}

\end{document}